\documentclass[sigconf]{acmart}
\AtBeginDocument{%
  }

\copyrightyear{2025}
\acmYear{2025}
\setcopyright{acmlicensed}\acmConference[SIGSPATIAL '25]{The 33rd ACM International Conference on Advances in Geographic Information Systems}{November 3--6, 2025}{Minneapolis, MN, USA}
\acmBooktitle{The 33rd ACM International Conference on Advances in Geographic Information Systems (SIGSPATIAL '25), November 3--6, 2025, Minneapolis, MN, USA}
\acmDOI{10.1145/3748636.3763223}
\acmISBN{979-8-4007-2086-4/2025/11}

\usepackage{enumitem}
\usepackage{tcolorbox}
\tcbuselibrary{listingsutf8}
\usepackage{listings}

\usepackage{multirow}
\usepackage{booktabs}
\usepackage{amsmath}
\usepackage{graphicx}
\usepackage[table,xcdraw]{xcolor}

\usepackage{tabularx}
\usepackage{array}
\usepackage{booktabs}
\usepackage{makecell}
\usepackage{hyperref}

\usepackage{caption}
\definecolor{lightblueframe}{HTML}{99CCFF}
\begin{document}



\title{TyphoFormer: Language-Augmented Transformer for Accurate Typhoon Track Forecasting}



\author{Lincan Li}
\affiliation{%
  \institution{Florida State University}
  \orcid{https://orcid.org/0000-0003-3797-4055}
  \city{Tallahassee}
  \country{USA}
}

\author{Eren Erman Ozguven}
\affiliation{%
 \institution{FAMU-FSU College of Engineering}
 \orcid{0000-0001-6006-7635}
 \city{Tallahassee}
 \country{USA}}

\author{Yue Zhao}
\affiliation{%
  \institution{University of Southern California}
  \orcid{https://orcid.org/0000-0003-3401-4921}
  \city{Los Angeles}
  \country{USA}}

\author{Guang Wang}
\affiliation{%
  \institution{Florida State University}
  \orcid{https://orcid.org/0000-0002-7739-7945}
  \city{Tallahassee}
  \country{USA}
}

\author{Yiqun Xie}
\affiliation{%
 \institution{University of Maryland}
 \orcid{https://orcid.org/0000-0002-6439-1333}
 \city{College Park}
 \country{USA}}

\author{Yushun Dong}
\affiliation{%
  \institution{Florida State University}
  \orcid{https://orcid.org/0000-0001-7504-6159}
  \city{Tallahassee}
  \country{USA}}
  
\renewcommand{\shortauthors}{Lincan Li et al.}

\begin{abstract}
Accurate typhoon track forecasting is crucial for early system warning and disaster response. While Transformer-based models have demonstrated strong performance in modeling the temporal dynamics of dense trajectories of humans and vehicles in smart cities, they usually lack access to broader contextual knowledge that enhances the forecasting reliability of sparse meteorological trajectories, such as typhoon tracks. To address this challenge, we propose \href{https://github.com/LabRAI/TyphoFormer}{\textbf{TyphoFormer}}, a novel framework that incorporates natural language descriptions as auxiliary prompts to improve typhoon trajectory forecasting. For each time step, we use Large Language Model (LLM) to generate concise textual descriptions based on the numerical attributes recorded in the North Atlantic hurricane database. The language descriptions capture high-level meteorological semantics and are embedded as auxiliary special tokens prepended to the numerical time series input. By integrating both textual and sequential information within a unified Transformer encoder, TyphoFormer enables the model to leverage contextual cues that are otherwise inaccessible through numerical features alone. Extensive experiments are conducted on HURDAT2 benchmark, results show that our TyphoFormer consistently outperforms other state-of-the-art baseline methods, particularly under challenging scenarios involving nonlinear path shifts and limited historical observations.
\end{abstract}




\maketitle

\section{Introduction}
Typhoons and hurricanes pose significant threats to coastal regions, causing devastating damage to infrastructure, ecosystems, and human lives~\cite{ramachandran2025addressing,wachnicka2020major}. The southeastern United States, especially Florida, is facing increasingly frequent and intense storms~\cite{oliver2019climate,malmstadt2009florida}. For instance, Hurricane IAN in 2022 resulted in over 150 fatalities and an estimated \$113 billion in economic losses, making it one of the costliest hurricanes in U.S. history~\cite{wang2024impacts,bushong2023critical}. With the rise of sea surface temperatures and the evolution of climate patterns, accurate typhoon trajectory forecasting has become more critical than ever for emergency planning, resource allocation, and early evacuation~\cite{wang2024changes}. Reliable predictions of storm movement not only support disaster mitigation efforts but also enable authorities to issue timely warnings, minimizing potential casualties and disruption~\cite{yu2022impact}. In light of the substantial socio-economic impact of such extreme weather events, there is a growing demand for advanced, data-driven forecasting methods that can complement or enhance traditional meteorological models.

Existing approaches for typhoon track forecasting generally fall into two main categories: (I) \textbf{time series modeling methods} and (II) \textbf{physics-based numerical models}. To address the need for data-driven forecasting, a growing body of work has explored modeling typhoon tracks as temporal sequences using deep learning techniques~\cite{wang2024multimodal}. These methods treat the storm's historical positions and attributes as a time series and aim to predict its future trajectory based on learned temporal patterns using deep learning-based methods. Early approaches employed recurrent neural network (RNN)-based architectures such as LSTM, GRU, and their variants to capture sequential dependencies in storm movement~\cite{song2022novel,wei2020nearshore}. More recent advances have leveraged Transformer-based architectures, which offer improved long-range modeling capabilities and greater scalability for sequence forecasting~\cite{zhou2021informer,wu2021autoformer}. Methods such as InFormer~\cite{zhou2021informer} and AutoFormer~\cite{wu2021autoformer} have demonstrated promising results in related time series domains, particularly for long-horizon prediction tasks. While these models excel in learning from past trajectory data, they lack an awareness of external contextual information, such as meteorological semantics or high-level event dynamics, that may further enhance predictive accuracy.



As a basic tool in traditional meteorological analysis, physics-based numerical models are also widely used in operational typhoon forecasting~\cite{iwasaki1987performance,knaff2007statistical}. Numerical models simulate atmospheric dynamics by solving complex partial differential equations derived from physical laws~\cite{bauer2015quiet}, such as the Navier–Stokes Equations~\cite{rozanova2010typhoon} and Boundary Layer Structure Modeling~\cite{kepert2010slab}. Prominent forecasting systems, including the the Global Forecast System (GFS)~\cite{halperin2020comparison} and Climatology and Persistence Model~\cite{knaff2007statistical}, use large-scale initial conditions, such as pressure fields, wind vectors, and sea surface temperatures, to estimate future storm trajectories. While these models offer physical interpretability and have been refined through decades of meteorological research, they require gigantic computational resources (e.g., supercomputing center) and are sensitive to initial condition uncertainties. Furthermore, their performance may degrade in data-sparse regions or during rapidly evolving storm phases, where high-resolution input data are limited or delayed. As a result, complementary deep learning and data-driven techniques have been increasingly explored to address these limitations inherent in traditional approaches.

Despite the aforementioned significant advancements, research on typhoon track prediction still faces several critical challenges. First, existing time series models often rely solely on numerical input features and fail to incorporate higher-level semantic knowledge that could enhance forecast reasoning. Second, most approaches lack interpretability and contextual awareness, limiting their robustness in complex or data-sparse scenarios. Third, physics-based models, while grounded in atmospheric theory, are computationally expensive and sensitive to initialization errors, making them difficult to scale in real-time applications. Different from existing works, we propose a novel framework that integrates natural language prompts generated from numerical meteorological data into a Transformer-based forecasting model. By embedding these LLM-derived textual descriptions as auxiliary tokens, our approach provides rich contextual guidance that complements the numerical trajectory input and improves prediction accuracy under uncertainty. Our contributions are summarized as follows:
\begin{itemize}[leftmargin=*,itemsep=0pt]
\item A natural language-augmented Transformer framework is proposed for accurate typhoon track forecasting, which integrates LLM-generated textual descriptions into the modeling of sparse spatial-temporal typhoon trajectories. 

\item We design PGF mechanism to realize effective embed then fuse the LLM-based textual prompts with meteorological records, enabling the model to incorporate semantic prior knowledge.

\item Detailed experiments on HURDAT2 dataset demonstrate that our TyphoFormer\footnote{https://github.com/LabRAI/TyphoFormer} achieves improved accuracy over state-of-the-art baseline methods, especially in challenging prediction scenarios.

\end{itemize}

\section{Methodology}

\begin{figure}[th]
    \centering
\includegraphics[width=0.99\columnwidth]{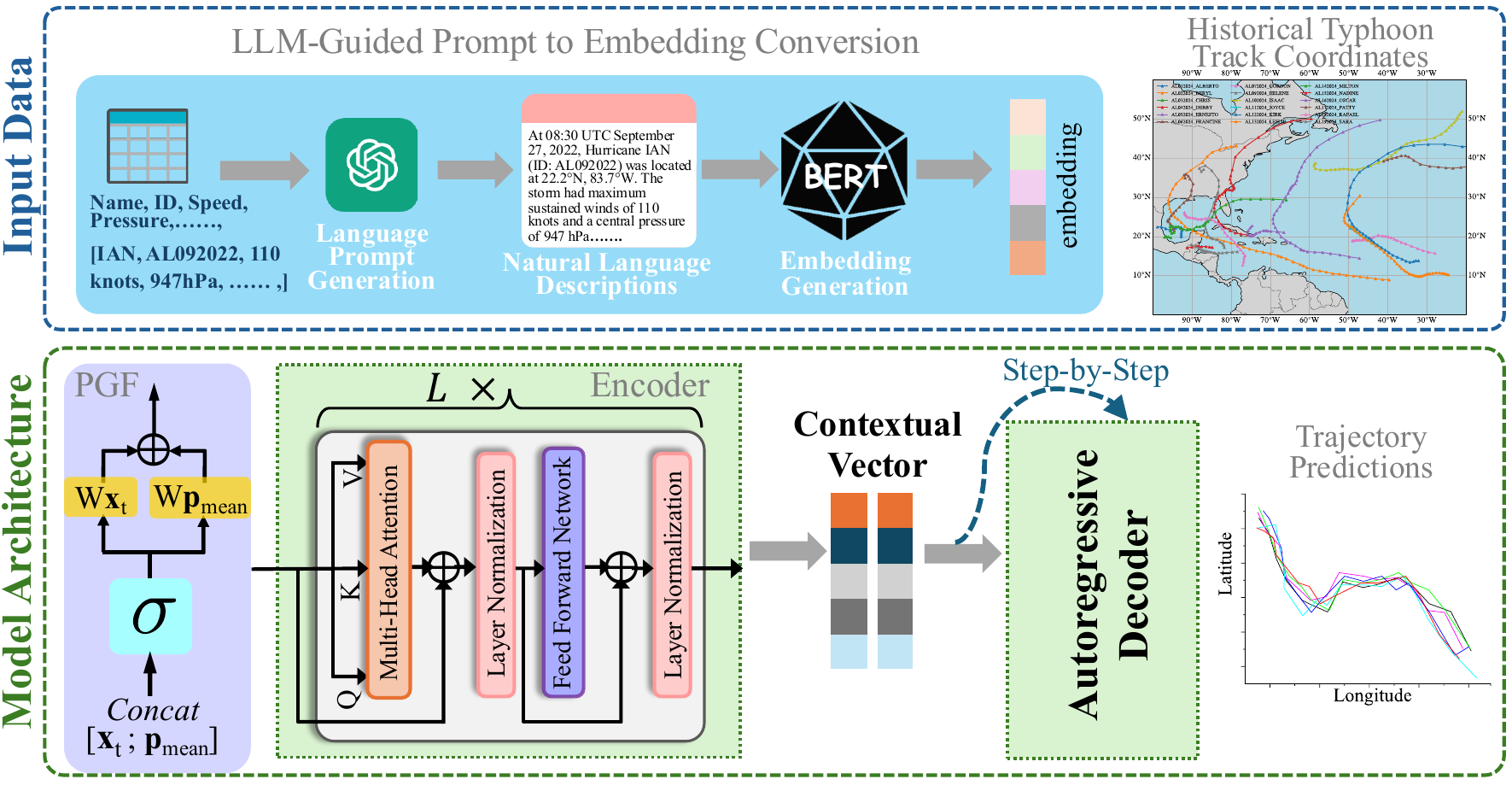}
    \caption{The overall model architecture of TyphoFormer.}
\label{model-architecture}
\end{figure}

We encode both structured meteorological features and textual descriptions into a unified Transformer-based architecture, enabling the model to capture both temporal dynamics and contextual knowledge. The overall model architecture is illustrated in Figure~\ref{model-architecture}.

\subsection{Trajectory Data and Problem Formulation}

We formulate typhoon track forecasting as a sequence prediction task over spatiotemporal meteorological data. Specifically, each typhoon trajectory is represented as a multivariate time series, where each time step includes numerical features such as timestamp, latitude, longitude, maximum sustained wind speed, and central pressure. Let a single typhoon trajectory be represented as a sequence of records: $\mathcal{X} = \{\mathbf{x_1}, \mathbf{x_2}, \dots, \mathbf{x_T}\}$, where $\mathbf{x_t} \in \mathbb{R}^d$ denotes the meteorological feature vector at time step $t$, and $d$ is the feature dimension. Our objective is to predict the next geographical position (e.g., latitude and longitude) of the typhoon at $T+1$ step, given the historical observations. This task can be formulated as: $\hat{y}_{T+1} = f_{\phi}(\mathbf{x_1}, \mathbf{x_2},\dots,\mathbf{x_T})$, where $f_{\phi}(\cdot)$ is the forecasting model, and $\hat{y}_{T+1} \in \mathbb{R}^2$ is the predicted coordinates.

\subsection{Language Contextual Prompt Generation}
To enhance the model's contextual understanding of meteorological dynamics, we generate natural language descriptions (referred to as semantic prompts) for each time step in the trajectory data. Unlike conventional approaches that rely solely on structured numerical inputs, our method introduces a textual layer of information that captures the same data in a human-readable, semantically coherent format. These prompts are automatically generated using a GPT-4o through API-based batch inference. Given a single line record from the trajectory dataset, such as:
\begin{verbatim}
AL142024_MILTON,20241010,0030,L,HU,27.4N,82.6W,100,
958,180,170,110,220,60,60,70,90,30,30,30,30,20
\end{verbatim}

\noindent which corresponds to a position in Hurricane MILTON at 00:30 UTC on 10/10/2024, the following natural language prompt is produced:
\begin{tcolorbox}[colback=gray!5!white, colframe=lightblueframe, fontupper=\footnotesize, left=1pt, right=1pt, top=1pt, bottom=1pt, title=LLM-generated contextual prompt]
At 00:30 UTC on October 10, 2024, Hurricane MILTON (AL142024) was located at 27.4°N, 82.6°W. The storm had maximum sustained winds of 100 knots and a central pressure of 958 hPa. The radius of 34-knot winds extended 180 km in the northeast quadrant, 170 km in the southeast, 110 km in the southwest, and 220 km in the northwest. Meanwhile, the radius of 50-knot winds reached 60 km in both the northeast and southeast, and 70 km and 90 km in the southwest and northwest quadrants, respectively. For 64-knot hurricane-force winds, the extent was 30 km in all four quadrants. The radius of the eye was estimated at 20 km.
\end{tcolorbox}

\noindent These natural language prompts encode meteorological knowledge in a compact form that would otherwise require explicit domain understanding. By providing a semantic summary of each state, they serve as high-level guidance to assist the forecasting model in interpreting patterns and trends within the trajectory. All prompts are generated offline using the GPT-4o API and cached for training and evaluation. This step introduces minimal computational overhead and significantly enriches the input representation.

\subsection{Text Embedding and Feature Fusion}

To integrate semantically rich natural language prompts into the forecasting model, we first encode each textual description into a dense sequence of embeddings using a pretrained language encoder. Specifically, we employ BERT to convert each prompt into a sequence $\mathcal{P} = \{\mathbf{p}_1, \mathbf{p}_2, \dots, \mathbf{p}_M\}$, where each token embedding $\mathbf{p}_m \in \mathbb{R}^d$. Meanwhile, the structured time series input is denoted as $\mathcal{X} = \{ \mathbf{x}_1, \mathbf{x}_2,\dots,\mathbf{x}_T \}$, with each $\mathbf{x}_t \in \mathbb{R}^d$ representing multivariate meteorological features at time step $t$.

Instead of directly concatenating these two modalities, we introduce a \textbf{\underline{P}rompt-aware \underline{G}ating \underline{F}usion} (PGF) mechanism to adaptively fuse the prompt information into the temporal sequence. First, we compute a condensed prompt representation by averaging the token embeddings:$\mathbf{p}_{\text{mean}} = \frac{1}{M} \sum_{m=1}^{M} \mathbf{p}_m$. Then, for each time step $t$, we compute a gated fusion vector:
\begin{equation}
\tilde{\mathbf{x}}_t = \sigma(W_g [\mathbf{x}_t; \mathbf{p}_{\text{mean}}] + b_g) \odot \mathbf{x}_t + \left(1 - \sigma(\cdot)\right) \odot \mathbf{p}_{\text{mean}}
\label{eq3-3}
\end{equation}

\noindent where $W_g \in \mathbb{R}^{2d \times d}$, $b_g \in \mathbb{R}^d$, and $\sigma$ is the sigmoid function. The fused input sequence becomes: $\mathcal{Z} = \{ \tilde{\mathbf{x}}_1, \tilde{\mathbf{x}}_2, \dots, \tilde{\mathbf{x}}_T \}$.

\noindent $\mathcal{Z} \in \mathbb{R}^{T \times d}$ is then fed into a Transformer encoder, which learns long-range dependencies and temporal patterns. PGF mechanism allows the model to selectively modulate the influence of the prompt across different time steps, depending on local signal strength and relevance. Compared to naive concatenation, PGF facilitates more controlled and informative integration of textual context, enhancing the model's ability to reason over sparse or ambiguous trajectories.

\subsection{Forecasting Head and Training}

After encoding the fused input sequence with the Transformer encoder, we obtain a sequence of contextualized representations $\{\mathbf{h}_1,\mathbf{h}_2, \dots, \mathbf{h}_T\}$. We then employ an Autoregressive decoder to generate future trajectory coordinates iteratively. At each decoding step, the input consists of the previous predicted output concatenated with $\mathbf{h}_T$, allowing the decoder to leverage both context and short-term dependencies. The model is trained by minimizing the Mean Squared Error (MSE) between predicted and ground-truth coordinates. Formally, it learns a mapping $f: \mathbf{h}_T \rightarrow \hat{\mathbf{y}}_{T+1}$, where $\hat{\mathbf{y}}_{T+1}$ is the predicted latitude and longitude. 

\section{Experiments}
\noindent \textbf{Dataset.} Experiments are carried on the well-known HURDAT2 dataset, which is the official North Atlantic hurricane database maintained by the U.S. National Hurricane Center. It provides detailed 6-hour records of tropical cyclones, including 22 critical meteorological features. Each typhoon event is recorded as a chronological sequence. We use the records from year 2004 to 2021 for training, while the data samples from 2022 to 2024 are used for testing.

\noindent \textbf{Evaluation metrics.} We use the commonly adopted prediction criteria \textit{Mean Absolute Error} (MAE) and a task-specific criteria--\textit{Spherical Distance Erro}r ($\Delta \text{R}$). $\Delta \text{R}$ measures the geospatial distance between the prediction and the ground-truth trajectories, which is calculated as: $\Delta \text{R}=\Delta \text{Re}\cdot \text{arccos} (\sin \varphi_p \sin \varphi_r+\cos \varphi_p \cos \varphi_r \cos(\lambda_p - \lambda_r))$, where $\text{Re} =6371km$ is the Earth's radius, $(\varphi_p,\lambda_p)$ and $(\varphi_r, \lambda_r)$ are the predicted / ground-truth latitude and longitude, respectively.

\noindent \textbf{Baseline methods.} We compare our TyphoFormer\footnote{https://github.com/LabRAI/TyphoFormer} against two categories of baselines. The first category is the classical numerical equation-based models, we choose the representative (I) CLIPER (Climatology and Persistence Model)~\cite{knaff2007statistical}, which extrapolates storm trajectories using historical patterns and persistence assumptions. The second category is time-series forecasting models, including recurrent models like (III) LSTM and (IV) GRU, as well as Transformer-based or MLP-based advanced architectures like (V) Informer~\cite{zhou2021informer}, (VI) Autoformer~\cite{wu2021autoformer}, and (VII) TSMixer~\cite{ekambaram2023tsmixer}. These baselines provide a diverse comparison across classical statistical models and advanced deep learning frameworks.

\subsection{Main Results}

\begin{table}[!htbp]
\centering
\footnotesize
\caption{Performance Comparison of Various Methods.}
\scriptsize The best values are in \textbf{bold font}, the second-best values are indicated with \underline{underlining}.
\vspace{-3pt}
\raggedright
\label{tab:comparison}
\setlength{\tabcolsep}{2pt}
\resizebox{\linewidth}{!}{
\begin{tabular}{lcccccccc}
\toprule
\multirow{2}{*}{\textbf{Models}} & \multicolumn{4}{c}{\textbf{MAE (All)}} & \multicolumn{4}{c}{\textbf{$\Delta$R (km) (All)}} \\
\cmidrule(lr){2-5} \cmidrule(lr){6-9}
 & 6h & 12h & 18h & 24h & 6h & 12h & 18h & 24h \\
\midrule
CLIPER & 0.235 & \underline{0.275} & 0.310 & 0.368 & \underline{34.265} & \underline{42.205} & 51.632 & \underline{58.268} \\
GRU & 0.367 & 0.405 & 0.493 & 0.640 & 50.480 & 69.397 & 90.875 & 103.894 \\
LSTM & 0.392 & 0.431 & 0.583 & 0.828 & 46.096 & 71.365 & 95.412 & 112.663 \\
Informer & 0.289 & 0.318 & 0.392 & 0.483 & 37.592 & 46.435 & 56.433 & 76.684 \\
Autoformer & 0.263 & 0.286 & 0.357 & 0.433 & 39.836 & 47.183 & 63.775 & 70.862 \\
TSMixer & \underline{0.214} & 0.268 & \underline{0.297} & \underline{0.353} & 35.720 & 45.265 & \underline{50.330} & 62.910 \\
\hline
TyphoFormer & \textbf{0.188} & \textbf{0.242} & \textbf{0.261} & \textbf{0.312} & \textbf{31.539} & \textbf{38.084} & \textbf{42.435} & \textbf{49.562} \\
\midrule
\multirow{2}{*}{\textbf{Models}} & \multicolumn{4}{c}{\textbf{MAE (2024)}} & \multicolumn{4}{c}{\textbf{$\Delta$R (km) (2024)}} \\
\cmidrule(lr){2-5} \cmidrule(lr){6-9}
 & 6h & 12h & 18h & 24h & 6h & 12h & 18h & 24h \\
\midrule
CLIPER & \underline{0.221} & {0.269} & \underline{0.294} & {0.361} & \underline{33.915} & {44.542} & \underline{48.580} & \underline{57.642} \\
GRU & 0.359 & 0.394 & 0.491 & 0.632 & 49.674 & 70.357 & 92.542 & 105.114 \\
LSTM & 0.388 & 0.422 & 0.587 & 0.835 & 44.203 & 69.051 & 94.555 & 109.108 \\
Informer & 0.290 & 0.316 & 0.395 & 0.487 & 37.843 & 48.791 & 57.445 & 78.497 \\
Autoformer & 0.257 & 0.284 & 0.358 & 0.435 & 39.128 & 46.085 & 65.039 & 72.827 \\
TSMixer & 0.228 & \underline{0.265} & 0.305 & \underline{0.347} & {35.148} & \underline{43.935} & 52.501 & 63.118 \\
\hline
TyphoFormer & \textbf{0.185} & \textbf{0.241} & \textbf{0.263} & \textbf{0.317} & \textbf{31.274} & \textbf{37.934} & \textbf{42.973} & \textbf{50.881} \\
\bottomrule
\end{tabular}
}
\end{table}

\begin{figure}[htbp]
\centering
\includegraphics[width=0.99\columnwidth]{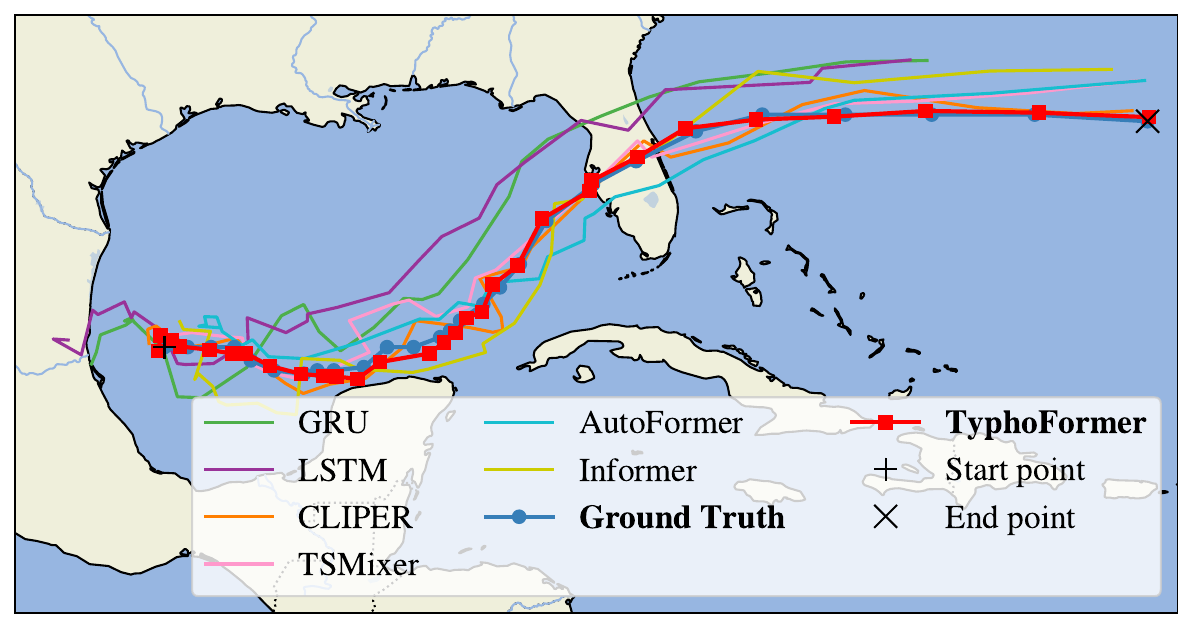}
\caption{Trajectory forecasting performance among all the methods on hurricane MILTON 2024.}
\label{exp-results-visualize}
\end{figure}

We report the performance of TyphoFormer and various baseline methods on typhoon track prediction in Table~\ref{tab:comparison}, using both Mean Absolute Error (MAE) and great-circle distance error ($\Delta$R) across four prediction horizons (6h, 12h, 18h, 24h). Results are reported on the full dataset (top) and on the 2024 subset (bottom). Overall, TyphoFormer consistently outperforms all baselines across all metrics and time horizons. Compared with the classical statistical model CLIPER and numerical forecast proxy NCEP, our model yields significantly lower MAE and $\Delta$R, especially at longer horizons (e.g., achieving a 24h MAE of 0.312 and $\Delta$R of 49.56 km vs. CLIPER's 0.368 and 58.27 km, respectively). This demonstrates the advantage of incorporating contextual knowledge through natural language-guided fusion. Among the deep learning baselines, TyphoFormer surpasses both recurrent models (GRU, LSTM) and Transformer-based architectures (Informer, Autoformer, TSMixer). Notably, our model shows the most stable performance across different forecast steps, with minimal degradation over longer horizons. For instance, the 24h $\Delta$R of TyphoFormer (49.56 km) is substantially lower than that of Informer (76.68 km) and Autoformer (70.86 km), indicating better spatial generalization. The improvements persist on the 2024 test year, confirming that TyphoFormer generalizes well to recent, unseen typhoon events. These results highlight the effectiveness of our method in leveraging language-augmented prior knowledge for robust typhoon trajectory forecasting.

\subsection{Case study on Hurricane MILTON}
Hurricane MILTON is developed in the Gulf of Mexico and made landfall in Florida in 2024 (see Figure~\ref{exp-results-visualize}). Notably, TyphoFormer achieves superior spatial fidelity across the entire storm path, especially in three critical segments where many baselines exhibit significant deviations: \textbf{(1) Early recurvature in the Gulf of Mexico:} In the initial west-to-east drift across the central Gulf, TyphoFormer correctly anticipates the cyclone's gentle northward bend around longitude 90°W, closely aligning with the observed dynamics. In contrast, Informer and CLIPER incorrectly maintain a more zonal trajectory, overshooting westward before eventually turning north. This suggests that TyphoFormer benefits from prompt-informed inductive priors that enhance recognition of subtle early steering shifts. \textbf{(2) Florida landfall and curvature east of 85°W:} The most critical test of trajectory fidelity occurs near the landfall point. TyphoFormer precisely captures both the timing and curvature of the landfall arc over Florida's western coast. In contrast, LSTM and GRU underestimate the curvature, producing flatter tracks that result in eastward positional bias post-landfall. These models tend to smooth transitions excessively, failing to model the nonlinear influence of land-sea friction and subtropical ridge interactions. \textbf{(3) Post-landfall drift into the Atlantic:} After crossing the Florida peninsula, TyphoFormer continues to trace the observed northeastward drift offshore with minimal divergence. Informer and Autoformer begin to diverge more drastically in this segment, due to error accumulation and limited ability to represent mesoscale dynamics once the storm enters the open Atlantic. TyphoFormer's attention-based structure with fused language conditioning contributes to its resilience in such dynamically uncertain regimes.


\section{Conclusions}
In this work, we present a novel typhoon trajectory forecasting framework named TyphoFormer, which integrates natural language prompts with time series modeling. By leveraging LLMs to generate semantically informative descriptions from raw meteorological records, and fusing these textual embeddings with numerical trajectory inputs through a prompt-aware Transformer encoder, our approach introduces a new paradigm for enriching data-driven prediction with contextual knowledge. Extensive experiments on HURDAT2 benchmark demonstrate that TyphoFormer consistently outperforms other state-of-the-art baselines, particularly in long-range and complex track scenarios. The findings highlight the potential of language-enhanced forecasting models in improving the accuracy and interpretability of meteorological prediction tasks.

\begin{acks}
This work is supported in part by the start up grant and the FYAP grant program provided by Florida State University.
\end{acks}

\bibliographystyle{ACM-Reference-Format}
\bibliography{main-authordraft}





\end{document}